\documentclass[journal]{IEEEtran}
\usepackage{amsmath,amsfonts}
\usepackage{algorithmic}
\usepackage{array}
\usepackage{textcomp}
\usepackage{stfloats}
\usepackage{url}
\usepackage{verbatim}
\usepackage{graphicx}
\usepackage[tight]{subfigure}
\def\BibTeX{{\rm B\kern-.05em{\sc i\kern-.025em b}\kern-.08em
    T\kern-.1667em\lower.7ex\hbox{E}\kern-.125emX}}
\usepackage{balance}

\usepackage{diagbox}
\usepackage{algorithm}
\usepackage{algorithmic}
\usepackage{mathtools}
\usepackage{bm}
\usepackage{soul}
\soulregister\cite7 
\soulregister\citep7 
\soulregister\citet7 
\soulregister\ref7 
\soulregister\pageref7 
\usepackage{booktabs}
\usepackage{bbding}
\usepackage{subcaption}
\usepackage{makecell,multirow}
\usepackage{colortbl}
\usepackage{xcolor}
\definecolor{mygray}{gray}{.9}
\definecolor{mypink}{rgb}{.99,.91,.95}
\definecolor{mycyan}{cmyk}{.3,0,0,0}
\usepackage[pagebackref=true,breaklinks=true, colorlinks,bookmarks=false]{hyperref}
\UseRawInputEncoding
\newcommand{\fakeparagraph}[1]{\vspace{1mm}\noindent\textbf{#1}}

\begin{document}
\title{Language-Guided Grasp Detection with Coarse-to-Fine Learning for Robotic Manipulation}

\markboth{Journal of \LaTeX\ Class Files,~Vol.~18, No.~9, September~2025}%
{How to Use the IEEEtran \LaTeX \ Templates}

\author{Zebin Jiang\,$^{2\dagger}$, Tianle Jin\,$^{2\dagger}$, Xiangtong Yao\,$^{2}$, Alois Knoll\,$^{2}$,~\IEEEmembership{Fellow,~IEEE}, Hu Cao\,$^{1,2*}$,~\IEEEmembership{Member,~IEEE}
\thanks{$\dagger$ Equal contribution, $*$ Corresponding author.}
}
\maketitle

\begin{abstract} 
Grasping is one of the most fundamental challenging capabilities in robotic manipulation, especially in unstructured, cluttered, and semantically diverse environments. Recent researches have increasingly explored language-guided manipulation, where robots not only perceive the scene but also interpret task-relevant natural language instructions. However, existing language-conditioned grasping methods typically rely on shallow fusion strategies, leading to limited semantic grounding and weak alignment between linguistic intent and visual grasp reasoning.
In this work, we propose Language-Guided Grasp Detection (LGGD) with a coarse-to-fine learning paradigm for robotic manipulation. LGGD leverages CLIP-based visual and textual embeddings within a hierarchical cross-modal fusion pipeline, progressively injecting linguistic cues into the visual feature reconstruction process. This design enables fine-grained visual-semantic alignment and improves the feasibility of the predicted grasps with respect to task instructions. In addition, we introduce a language-conditioned dynamic convolution head (LDCH) that mixes multiple convolution experts based on sentence-level features, enabling instruction-adaptive coarse mask and grasp predictions. A final refinement module further enhances grasp consistency and robustness in complex scenes.
Experiments on the OCID-VLG and Grasp-Anything++ datasets show that LGGD surpasses existing language-guided grasping methods, exhibiting strong generalization to unseen objects and diverse language queries. Moreover, deployment on a real robotic platform demonstrates the practical effectiveness of our approach in executing accurate, instruction-conditioned grasp actions. \textit{The code will be released publicly upon acceptance.}
 
\end{abstract}


\section{Introduction}
\IEEEPARstart{R}{obotic} grasping serves as a fundamental yet capability for autonomous manipulation, underpinning tasks ranging from industrial logistics and manufacturing to household assistance and service robotics~\cite{cao2022efficient,7583659,doi:10.1126/scirobotics.aat6385,fu2024mobilealohalearningbimanual}. The ultimate objective of robotic grasping is to endow robots with the ability to reliably detect, localize, and physically grasp objects in an open-world setting, handling diverse objects, occlusions, and ambiguity arising from human intention. Despite extensive research, achieving robust grasping in realistic environments, especially when guided by natural language instructions, remains a significant challenge.

\begin{figure}[t!]
  \centering
  \includegraphics[width=\linewidth]{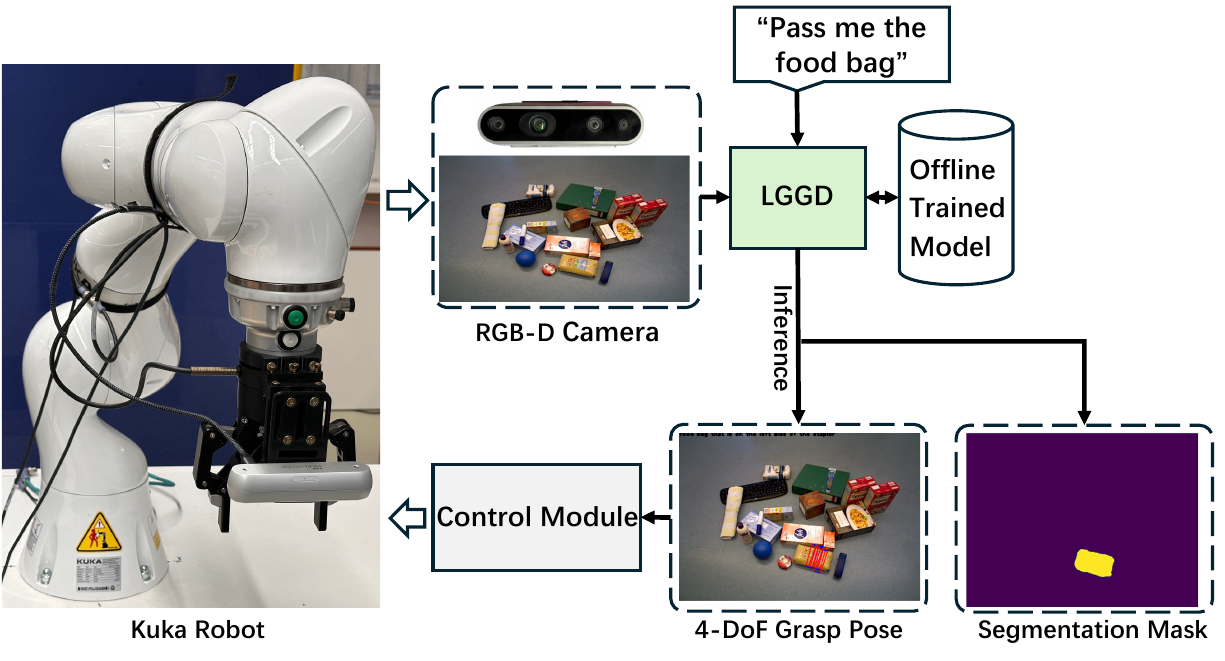}
  \caption{Overview of our system: An RGB-D camera mounted on the robot's wrist captures visual data of objects to be grasped. Our proposed LGGD generates 4-DoF grasp poses based on the RGB image and language query during the inference process. These generated grasp poses are then utilized by the control module to plan and execute robot trajectories for pick-and-place tasks.}
  \label{fig:abstract}
\end{figure}

Traditionally, analytical and model-based methods formed the first generation of grasp detection solutions, relying on classical mechanics, friction cone analysis, and 3D geometric modeling~\cite{1307170,844081}. These approaches assume precise object models and compute feasible contact points using force-closure or wrench-space formulations. While theoretically grounded, their applicability sharply declines when object models are incomplete or unavailable, or when objects are novel, deformable, or partially occluded-conditions that frequently occur in real domestic and industrial scenes. Moreover, collecting accurate mesh models for every object is impractical, and performance suffers when sensing noise, camera perspective distortion, or cluttered backgrounds are present.
The proliferation of deep learning triggered a paradigm shift from geometry-dependent methods to data-driven grasp inference~\cite{lenz2014deeplearningdetectingrobotic, mahler2017dex, 9787342, li2020event}. Discriminative grasping approaches conceptualize grasp detection as a selection problem: sample a pool of grasp hypotheses and evaluate them to select the best one. Two-stage methods, exemplified by Lenz \emph{et~al.}~\cite{lenz2014deeplearningdetectingrobotic} and GQCNN~\cite{mahler2017dex}, often produce high-quality grasp candidates but suffer from several limitations: (1) grasp candidate sampling is a bottleneck, (2) performance degrades when candidate coverage is insufficient, and (3) the separation of sampling and evaluation introduces error propagation. One-stage discriminative detectors~\cite{zhang2019real,kumra2017roboticgraspdetectionusing} merge hypothesis generation and scoring into a single process, improving computational efficiency, yet they still operate within a candidate-based paradigm and do not fundamentally exploit continuous per-pixel grasp reasoning.

This limitation motivated the rise of generative grasp detection, which reconceptualizes grasping as dense prediction over the image domain~\cite{morrison2018closinglooproboticgrasping}. Instead of evaluating discrete hypotheses, generative approaches directly output grasp quality, orientation, and gripper width per pixel. This conceptual leap eliminates the need for sampling, significantly improves inference speed, and harmonizes grasping with pixel-level perception pipelines such as semantic segmentation. Advancements including GR-ConvNet~\cite{kumra2020antipodal}, SE-enhanced designs~\cite{yu2022se}, ASPP-based receptive-field scaling~\cite{cao2021residual}, and Transformer-based perception~\cite{wang2022transformer,zhang2023htc} have all contributed to improving grasping robustness and accuracy under visual ambiguity. However, generative methods remain fundamentally \emph{purely visual}. They operate solely on image input and do not incorporate semantic intent or contextual constraints from natural language instructions.
The inability of visually driven grasping models to integrate task intent forms a crucial barrier to \mbox{human-robot} collaboration. Humans rarely specify grasps purely visually; instead, we express intentions linguistically:  \textit{
``pick up the red screwdriver''  ``grab the knife by the handle,'' ``pick up the leftmost apple,'' ``hand me the shallow bowl, not the deep one.''  }
Such statements encode object identity, attributes, relational placement, spatial disambiguation, safety semantics, and implicit operational constraints. A purely visual model has no access to this semantic layer of meaning and may inadvertently grasp a wrong object.

The emergence of large-scale vision-language models (VLMs), such as CLIP~\cite{radford2021learningtransferablevisualmodels}, BLIP~\cite{li2022blipbootstrappinglanguageimagepretraining}, and LLaVA~\cite{liu2023visualinstructiontuning}, has provided a powerful foundation for connecting linguistic reasoning with visual grounding. These models align language tokens with visual embeddings across broad conceptual spaces and have shown early promise in robot manipulation settings~\cite{shridhar2022cliport,vuong2024language,xu2023joint}. However, existing language-guided grasping systems primarily attach language to vision at a single stage of processing, either early (shared encoder input)~\cite{nguyen2024lightweight}, or late (final decision stage)~\cite{vo2024mask-guided, jin2024reasoninggrasping}. Such shallow or uni-directional fusion typically fails to maintain semantic context throughout the grasp inference pipeline, resulting in weak grounding of task intent, particularly for fine-grained grasping tasks (e.g., grasping a specific object part).

In this work, we address this gap by introducing \textbf{LGGD}, a coarse-to-fine language-guided grasp detection framework that performs multi-level vision-language fusion. Unlike prior approaches that fuse language only once, LGGD injects linguistic intent throughout the perception pipeline. First, the dual cross vision-language fusion module (DCVLF) enables symmetric exchange between visual tokens and word-level language features, allowing visual representations to attend to relevant linguistic cues while refining the language representation based on visual evidence. Second, we introduce a hierarchical language-guided upsampling module (LMAFN) that reconstructs high-resolution features while modulating multi-scale decoder activations using sentence-level semantics via Feature-wise Linear Modulation (FiLM) and attention, preserving instruction context during spatial detail recovery. Third, at the prediction stage, we employ a language-conditioned dynamic convolution head (LDCH) that forms instruction-conditioned decoding kernels through an expert-mixture formulation, improving semantic consistency for dense mask and grasp predictions. Finally, a lightweight residual refinement module corrects local artifacts and stabilizes grasp outputs in cluttered scenes. Another advantage of our approach is the training data: in addition to OCID-VLG~\cite{tziafas2023languageguidedrobotgraspingclipbased}, we train on the larger Grasp-Anything++~\cite{vuong2024language} dataset, which covers more object categories and scene variations with diverse natural language descriptions, improving robustness to novel objects, unusual phrasing, and out-of-distribution scenes. As shown in Figure~\ref{fig:abstract}, we deploy LGGD on a real robotic system for pick-and-place execution, consisting of a KUKA robot, a wrist-mounted RGB-D camera, the LGGD model, and a control module. In real-world grasping experiments, LGGD achieves reliable real-robot performance.

The key contributions of this work can be summarized as follows:

\begin{itemize}
    \item \textbf{Framework.} We propose LGGD, an end-to-end language-conditioned grasp detection framework that couples referring segmentation with dense grasp prediction in a coarse-to-fine manner.
    \item \textbf{Fusion + Decoding.} We introduce DCVLF for bidirectional word-level fusion, and a language-guided hierarchical upsampling module to recover spatial details under linguistic conditioning.
    \item \textbf{Instruction-specialized heads + Refinement.} We develop a language-conditioned dynamic convolution head (LDCH) for instruction-aware coarse predictions, together with residual refinement modules (RM) and multi-stage deep supervision to improve robustness in clutter.
    \item \textbf{Evaluation.} We validate our approach on OCID-VLG and Grasp-Anything++ and further demonstrate real-robot interactive pick-and-place performance under varying tabletop layouts.
\end{itemize}

\section{Related Work}

\subsection{Grasp Detection}
Grasp detection is a core research area in robotics, focused on enabling robots to recognize and perform object grasps in complex environments~\cite{ainetter2021end, depierre2018jacquard, shridhar2022cliport, vuong2023graspanything, dai2023graspnerf}. 

\fakeparagraph{Model-based Grasp Detection.}
Traditionally, grasp detection relies on analytical approaches~\cite{1307170, 844081}, which require 3D object models. The models are scanned and stored in a grasp pose database. During grasp pose detection, the object’s point cloud or 3D model data collected by the camera are used for object recognition and pose estimation. The object information is then matched in the database to retrieve the optimal grasp pose. This method has a relatively complex process involving object recognition, object pose estimation, and grasp pose generation. Its core is to construct a grasp contact model based on point contacts, analyzing contact forces and torques. It is suitable for simple scenarios with only a small number of objects but requires obtaining object models in advance and annotating grasp poses. It has poor generalization ability for unknown objects and does not fully utilize cameras or other sensors to enhance the robot's perception of the environment. In real robotic operation environments, it is often impossible to obtain precise models of all objects that need to be manipulated, and environmental interference can also affect performance. Therefore, model-based methods have limited applications in real-world scenarios.

\fakeparagraph{Discriminative Grasp Detection.}
With the emergence of deep learning, grasp detection witnessed substantial performance improvements. Lenz~\emph{et~al.}~\cite{lenz2014deeplearningdetectingrobotic} propose an early two-stage cascade, in which the first network rapidly filters out grasp candidates, while the second network conducts a more refined evaluation based on RGB, depth, and surface-normal features. Following this paradigm, Mahler \emph{et~al.}~\cite{mahler2017dex} introduce the Grasp Quality Convolutional Neural Network (GQCNN), which employs a Region Proposal Network (RPN) to sample grasp-relevant regions and subsequently evaluates them using a learned grasp-quality metric to determine the optimal grasp pose.
To overcome the inherent latency of multi-stage approaches, one-stage methods combine grasp proposal generation and scoring into a single forward pass, offering significantly improved efficiency. Zhang \emph{et~al.}~\cite{zhang2019real} modified the Fully Convolutional Network (FCN)~\cite{zhou2018fully} by replacing the blue channel of the RGB image with depth information, forming an RG-D input that consistently outperforms its RGB counterpart. Kumra \emph{et~al.}~\cite{kumra2017roboticgraspdetectionusing} further demonstrated the benefit of multimodal fusion by designing a dual-branch ResNet architecture that independently extracts RGB and depth features before combining them for grasp prediction.
Building on the one-stage paradigm, subsequent works introduced additional innovations. Xu \emph{et~al.}~\cite{xu2023instance} presented the Single-Shot Grasp (SSG) detector inspired by YOLACT~\cite{bolya2019yolactrealtimeinstancesegmentation}, integrating instance segmentation with grasp detection to enable instance-aware grasping. Xu \emph{et~al.}~\cite{xu2022gknet} eliminated reliance on predefined anchor boxes by proposing an \emph{anchor-free} grasp detector. Extending this framework, Wu \emph{et~al.}~\cite{wu2021real} reformulated angle estimation as a classification problem, whereas Chen \emph{et~al.}~\cite{cheng2023anchor} introduced a continuous angle-encoding scheme that resolves the discontinuity of discrete angle bins and yields more accurate angular predictions.

\fakeparagraph{Generative Grasp Detection.}
Formulating robotic grasping as a dense prediction problem was first demonstrated by Morrison \emph{et~al.}~\cite{morrison2018closinglooproboticgrasping} with the Generative Grasping Convolutional Neural Network (GG-CNN). Instead of generating and evaluating discrete grasp hypotheses, GG-CNN directly predicts a grasp pose at every pixel location, analogous to semantic segmentation. Owing to its compact and fully convolutional design, GG-CNN enables near-real-time inference and supports dynamic manipulation scenarios.
A major progression in generative grasping followed with the Generative Residual Convolutional Neural Network (GR-ConvNet) proposed by Kumra \emph{et~al.}~\cite{kumra2020antipodal}. GR-ConvNet incorporates residual learning and RGB–D fusion, compressing feature maps via five residual encoder blocks followed by symmetric transposed-convolution decoding. From the final decoder layer, the network produces four dense grasp maps that represent, respectively, the grasp quality, the orientation angle \(\cos 2\theta\) and \(\sin 2\theta\), and the required gripper width. This parameterization enables accurate and continuous grasp prediction while maintaining real-time execution speed.
To further enhance generative models, attention mechanisms have proven highly effective. Yu \emph{et~al.}~\cite{yu2022se} augmented a U-Net with squeeze-and-excitation blocks, enabling adaptive channel-wise feature reweighting and improved robustness in cluttered environments. To expand receptive fields, Cao \emph{et~al.} introduced atrous spatial pyramid pooling (ASPP)~\cite{cao2021residual}, and subsequently proposed gaussian-based grasp representation and the Multi-Dimensional Attention Fusion Network (MDAFN)~\cite{cao2022efficient}, which jointly applies spatial and channel attention to achieve multi-scale feature aggregation with minimal computational overhead.
Beyond CNN-based approaches, Wang \emph{et~al.}~\cite{wang2022transformer} incorporated hierarchical Swin-UNet~\cite{cao2022swin}, leveraging windowed self-attention for improved global context modelling. Zhang \emph{et~al.}~\cite{zhang2023htc} further advanced this direction by combining a Transformer encoder with a convolutional decoder via a Vision-Mamba architecture, unifying long-range reasoning with localized geometric detail and establishing state-of-the-art (SOTA) performance.
Despite these advances, existing generative grasping frameworks remain purely vision-based. They do not interpret or reason over linguistic commands, and thus cannot incorporate task-level semantic intent, such as grasping \textit{``the red mug by the handle'' or ``avoiding the sharp metal blade''.} Without explicit language grounding, models may detect feasible grasps but lack awareness of which object should be grasped and in what manner, limiting their utility in human-interactive and instruction-driven robotic systems.

\subsection{Language-Guided Grasp Detection}

Recent progress in robotic grasping has started to overcome the limitation of analytical and deep learning-based approaches through the incorporation of language understanding, allowing robots to execute grasps based on natural language guidance~\cite{lu2023vlgrasp6dofinteractivegrasp, sun2023languageguided, nguyen2024languagedriven6dofgraspdetection, 9811367, vuong2024language, nguyen2024lightweight, xu2023joint, 9812360}. 

Early language-driven systems convert the instruction into a referring object mask and subsequently run a conventional grasp detector on the masked crop. Shridhar \emph{et~al.} proposed CLIPort~\cite{shridhar2022cliport}, which uses CLIP embeddings to align the pixel‑wise feature map with the textual query, producing a dense relevance heat‑map that guides a Transporter network for planar pick-and-place manipulation. While CLIPort handles compositional commands, its two-stage structure, grounding followed by grasp prediction, means that grounding errors directly degrade the final grasp. To solve the problem of accumulated errors in the two-stage method, Liu \emph{et~al.}~\cite{liu2024hierarchical} introduce a hierarchical fusion network that integrates sentence-level, noun-level, and spatial-phrase cues with global, object-level, and spatial visual features extracted by CLIP.
Recent researches focus on the fusion of visual and language modalities. Chen et al.~\cite{9561994} used residual networks (ResNet)~\cite{he2016deep} to process visual information and long short-term memory networks (LSTM)~\cite{lstm} to process text input. However, due to the limitation of internal feature alignment, this method has limitations in processing complex instructions. The emergence of vision-language models (VLMs) provides new technical options for this task. Pre-trained backbones such as BERT~\cite{devlin2019bertpretrainingdeepbidirectional} and CLIP~\cite{radford2021learningtransferablevisualmodels} provide strong language and cross-modal representations. BERT captures contextual token dependencies through masked language modeling, whereas CLIP aligns image and text embeddings via a contrastive objective learned from 400 million image-text pairs. 

More recently, GraspMamba~\cite{nguyen2024graspmamba} introduces a Mamba-based language-driven grasp detection framework with hierarchical feature fusion, aiming to efficiently integrate text and multi-scale visual information to improve both accuracy and inference speed. While this linear-time architecture improves long-range dependency modeling and inference efficiency over diffusion- or transformer-based methods, its multimodal fusion mechanism still struggles to preserve fine-grained semantics through the full perception pipeline, making it difficult to accurately localize graspable regions when instructions are complex or object parts are visually similar. Meanwhile, Vo \emph{et~al.}~\cite{vo2024mask-guided} propose a mask-guided attention approach to inject language cues into the visual stream by restricting attention to segmentation-derived regions. Although this strategy enhances grounding within localized masks, the fusion is typically applied only at a single stage and relies on pre-extracted masks, limiting semantic propagation and weakening intent consistency. Similarly, segmentation-guided approaches such as GraspSAM~\cite{noh2024graspsam} convert language or clicking inputs into bounding-box prompts and inject them only at the decoder stage. As a result, all these fusion designs exhibit insufficient bidirectional interaction between modalities, resulting in suboptimal interpretation of fine-grained grasp intentions, particularly in cluttered or ambiguous scenes.

To better align grasp prediction with human intent, our method goes beyond simply using CLIP encoders. We embed language instructions throughout the whole perception pipeline by introducing a hierarchical cross-modal fusion mechanism, enabling bidirectional interaction between image and text tokens. This design preserves fine-grained semantic cues and ensures that visual features are progressively refined by linguistic intent at multiple scales. As a result, the detected grasp poses maintain strong semantic consistency with the user-specified command.

\begin{figure}[t!]
  \centering
  \includegraphics[width=0.6\linewidth]{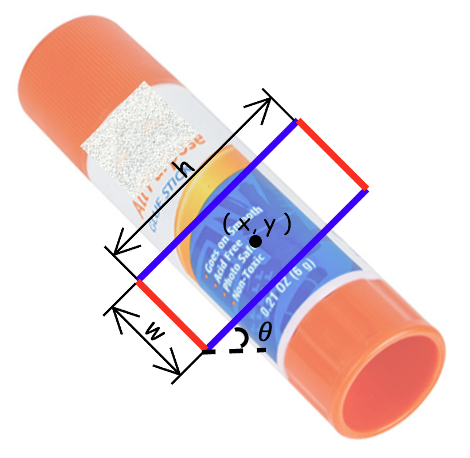}
  \caption{Rectangular grasp representation.}
  \label{fig:grasp}
\end{figure}

\section{Grasp Representation}
Given an RGB image and a text prompt specifying the target object, our goal is to detect a grasp pose in the image that corresponds to the provided description. To represent grasps, we adopt the widely used rectangle-based grasp formulation employed in prior works~\cite{ainetter2021end, depierre2018jacquard, cao2022efficient}, as illustrated in Figure~\ref{fig:grasp}. For an $N$-channel input image $\mathbf{I}\in\mathbb{R}^{H\times W\times N}$, a planar grasp is defined by a four-tuple of parameters:

\begin{equation}
  g=\{(x,y),\theta,w,q\}.
\end{equation}
where $(x,y)$ denotes the grasp center in image coordinates, $\theta$ is the rotation angle of the end-effector about the camera  $z$-axis, $w$ is the gripper opening width, and  $q\in[0,1]$ represents the estimated probability of grasp quality.

\begin{figure*}[t!]
  \centering
  \includegraphics[width=\linewidth]{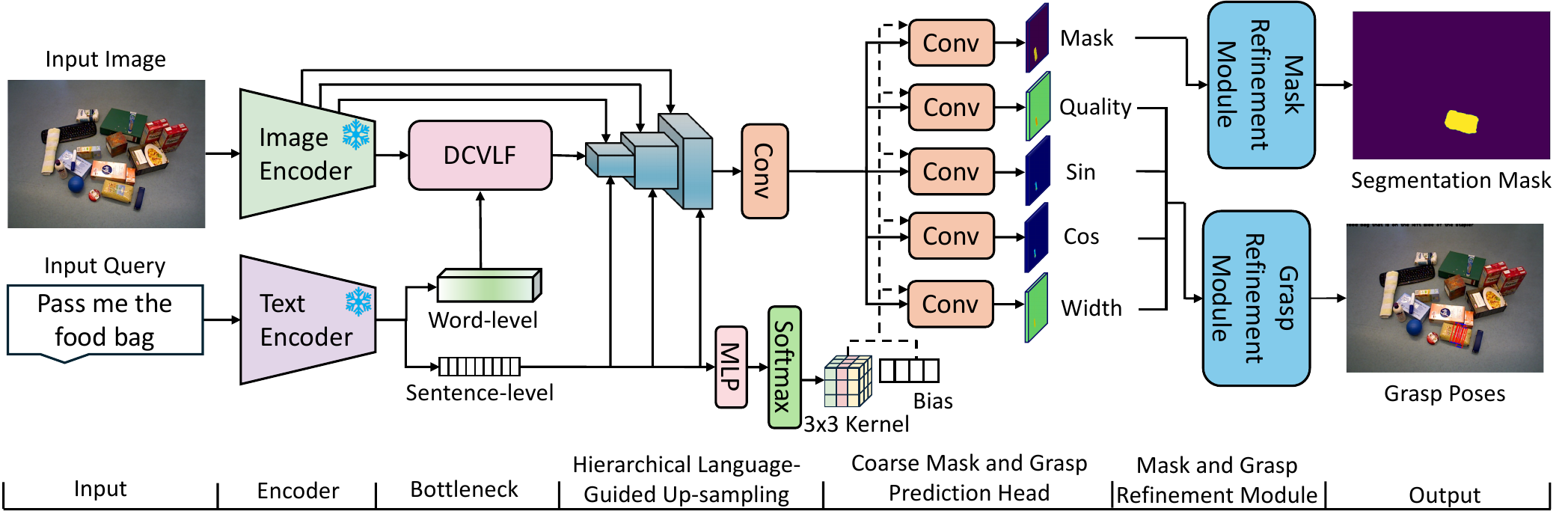}
  \caption{An overview of our proposed LGGD framework. Given an RGB image and a natural-language command, a CLIP-based image encoder and text encoder extract visual features and word/sentence embeddings. The Dual Cross Vision-Language Fusion (DCVLF) bottleneck aligns the two modalities, after which hierarchical language-guided upsampling progressively refines spatial details according to the textual intent. A coarse mask and grasp prediction head outputs the segmentation mask, grasp quality, angle, and gripper width. Finally, the mask refinement and grasp refinement modules sharpen boundaries and stabilize grasp poses, producing accurate, instruction-consistent grasp poses.
  }
  \label{fig:main}
\end{figure*}

\section{Method}

As illustrated in Figure~\ref{fig:main}, our language-guided grasp detection framework with coarse-to-fine learning (LGGD) consists of several key components: pretrained CLIP-based image and text encoders, a dual cross vision-language fusion (DCVLF) bottleneck, hierarchical language-guided upsampling modules, a coarse mask and grasp prediction head, and subsequent mask and grasp refinement modules.

\subsection{CLIP-based Encoders}
Contrastive Language-Image Pre-training (CLIP)~\cite{radford2021learningtransferablevisualmodels} represents a significant advancement in vision-language models that learn visual representations through natural language supervision rather than traditional supervised learning on manually annotated datasets. The CLIP model consists of two encoders: an image encoder and a text encoder that map visual and textual inputs into a shared embedding space. The image encoder processes RGB images through a deep network (commonly ResNet~\cite{he2016deep} or Vision Transformer~\cite{vaswani2023attentionneed} architectures), while the text encoder processes natural language descriptions through transformer-based architectures. Both encoders produce normalized feature vectors that enable direct comparison through cosine similarity. The training process utilizes contrastive learning on large-scale image-text pairs. Given a batch of $N$ image-text pairs, CLIP computes embeddings for all images $\{I_1, I_2, \ldots, I_N\}$ and texts $\{T_1, T_2, \ldots, T_N\}$. The image encoder generates normalized embeddings $i_k$ for each image $I_k$, while the text encoder produces normalized embeddings $t_k$ for each text $T_k$. 

\begin{figure*}[t!]
  \centering
  \includegraphics[width=0.9\linewidth]{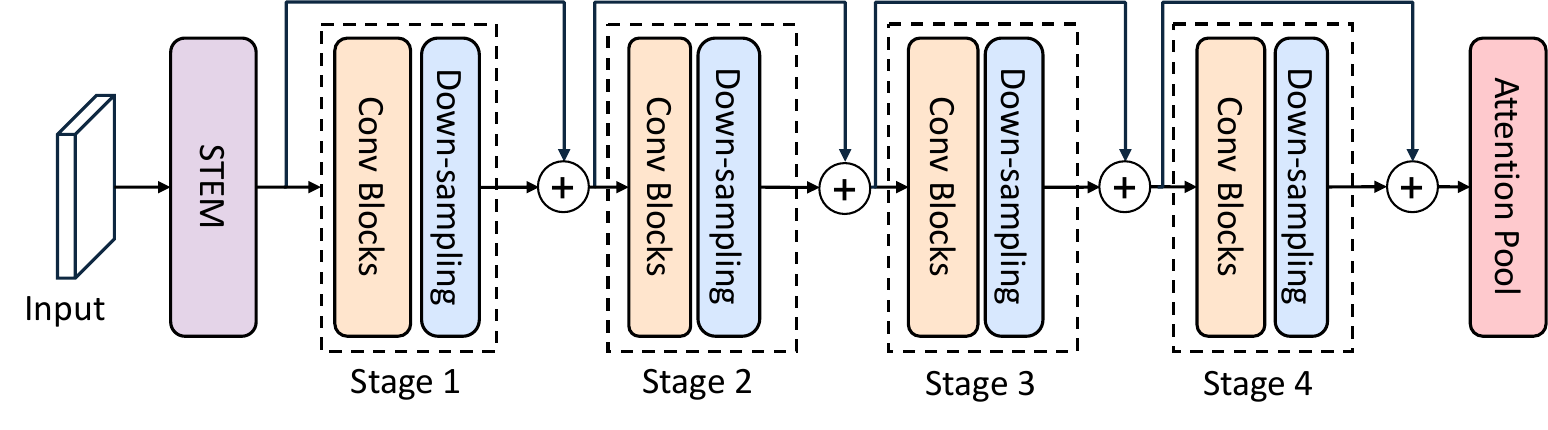}
  \caption{Structure of Image Encoder. The CLIP-based ResNet-50 backbone extracts progressively abstracted feature maps across four residual stages, while attention pooling enhances global semantic perception.}
  \label{fig:clip-image}
\end{figure*}

\fakeparagraph{Image Encoder.}
CLIP has established robust cross-modal feature correlations through large-scale pre-training on approximately 400 million image-text pairs. These correlations persist even when the model parameters are frozen during downstream task training. We employ a CLIP-based image encoder to utilize the pre-established semantic consistency between its visual and textual representations, preparing for subsequent feature fusion. Figure~\ref{fig:clip-image} illustrates the structure of the CLIP-based ResNet-50 image encoder. Given a resized RGB input image $I \in \mathbb{R}^{224 \times 224 \times 3}$, the CLIP-based ResNet-50 encoder processes it through a hierarchical sequence of convolutional blocks to extract features.
Initially, the image passes through the Stem block, followed by batch normalization and ReLU activations. This process reduces the spatial dimensions by half and outputs an initial feature map $x_{\text{stem}} \in \mathbb{R}^{112 \times 112 \times 64}$.


The output is then processed by four ResNet stages, each built from standard bottleneck blocks but preceded by an anti-alias (AA) down-sampling unit that performs average pooling $(2 \times 2, s = 2)$ followed by a stride-1 convolution. The whole process can be formulated as follows:

\begin{align}
x_0 &= f_{\text{stage1}}(x_{\text{stem}}), & x_0 &\in \mathbb{R}^{56\times56\times256},\\
x_1 &= f_{\text{stage2}}(x_0),            & x_1 &\in \mathbb{R}^{28\times28\times512},\\
x_2 &= f_{\text{stage3}}(x_1),            & x_2 &\in \mathbb{R}^{14\times14\times1024},\\
x_3 &= f_{\text{stage4}}(x_2),            & x_3 &\in \mathbb{R}^{7\times7\times2048},\\
x_4 &= f_{\text{attn}}(x_3),              & x_4 &\in \mathbb{R}^{7\times7\times2048}.
\end{align}

To enhance global spatial context modeling, we apply an attention pooling module to the stage-4 feature map $x_3$ and obtain an attention-enhanced representation $x_4$. The module leverages multi-head self-attention to aggregate information across the $7\times7$ spatial grid, enabling long-range interactions and improving the expressiveness of high-level visual features for subsequent multimodal fusion. In our design, the intermediate feature maps $\{x_0, x_1, x_2\}$ are retained as shallow features for skip connections in the hierarchical upsampling decoder, while $x_4$ is used as the visual input to the DCVLF bottleneck for vision-language interaction.


\fakeparagraph{Text Encoder.}
We employ the language branch of CLIP~\cite{radford2021learningtransferablevisualmodels} as our text encoder. It takes natural language prompts as input and first tokenizes them into 77 tokens using a Byte Pair Encoding (BPE) vocabulary ($\sim$49k tokens). Special tokens \texttt{<|startoftext|>} and \texttt{<|endoftext|>} are inserted at the beginning and end, respectively. Token IDs are embedded by adding word and learnable positional embeddings, which are then processed by a 12-layer Transformer encoder with a hidden dimension of 512.
The encoder produces two different granularities of text representations: $y_{\text{word}} \in \mathbb{R}^{77\times512}$ and $y_{\text{sentence}} \in \mathbb{R}^{512}$. $y_{\text{word}}$ corresponds to word-level contextual features and $y_{\text{sentence}}$ is derived from the final \texttt{<|endoftext|>} token followed by projection and $\ell_2$-normalization.
We employ the two types of features for different roles within our multi-modal design. Word-level features preserve fine-grained semantics such as spatial descriptions or object attributes (e.g. ``left,'' ``red,'' ``cup''). They are used in the DCVLF module, where visual features serve as queries and can dynamically attend to the most relevant tokens to achieve pixel-level grounding through bidirectional cross-attention. The sentence-level features provide a global semantic representation of the entire instruction. We use it as a high-level control signal in later up-sampling and coarse mask and grasp prediction head stages. During up-sampling, we use sentence-level features to modulate visual feature channels via Feature-wise Linear Modulation (FiLM)~\cite{perez2018film} to enforce task constraints (e.g., emphasizing the ``cup'' while suppressing irrelevant regions). In prediction head, we use sentence-level features to generate customized convolutional kernels, enabling task-specific behavior depending on the instruction. Through this design, our model can both align visual features with local linguistic concepts and adjust global processing according to the whole intent, leading to precise and instruction-consistent grasp generation.



\begin{figure*}[t!]
  \centering
  \includegraphics[width=0.9\linewidth]{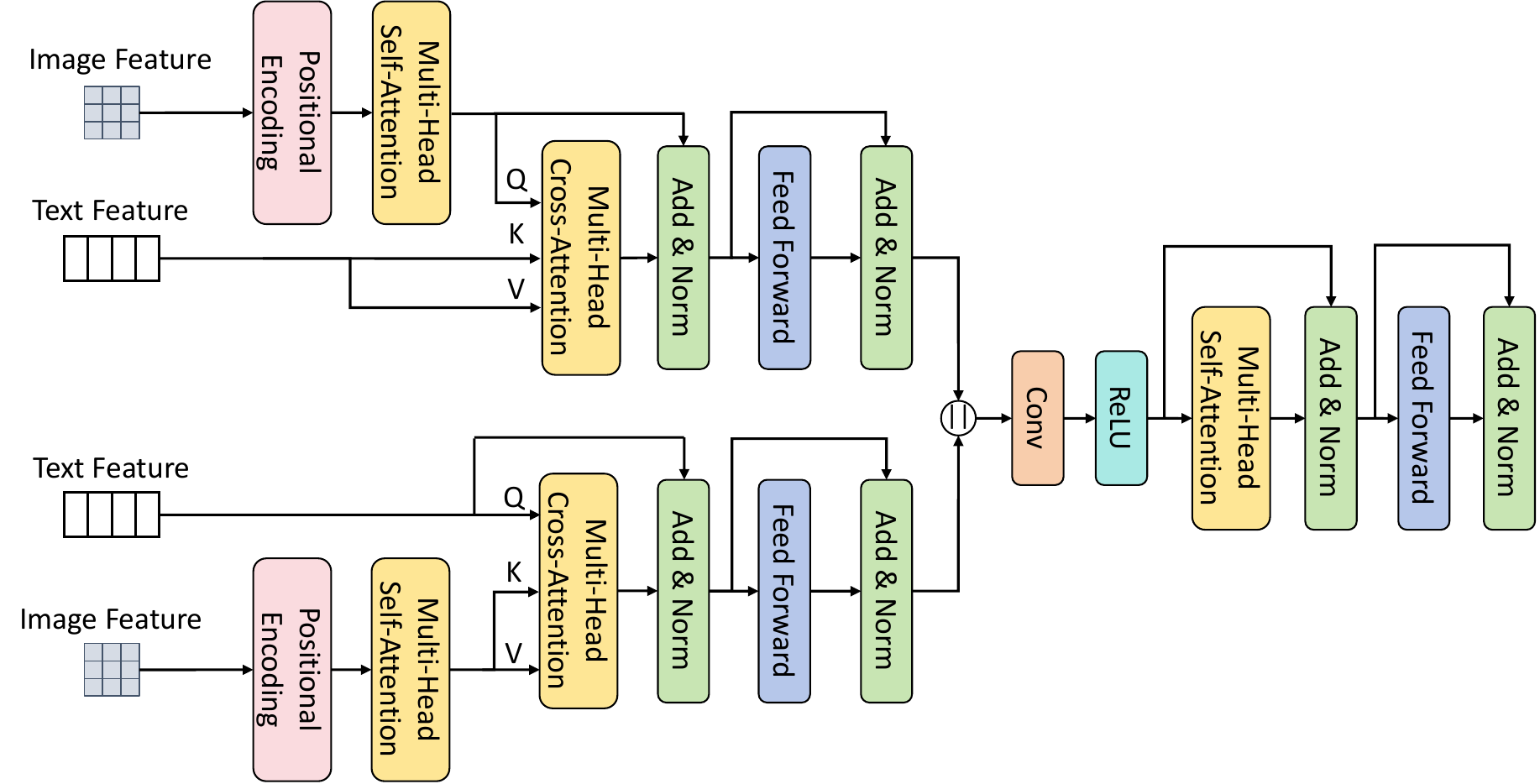}
  \caption{Structure of the DCVLF. Visual features are enriched with positional encoding and refined by a self-attention block. Bidirectional cross-attention then allows image features to attend to language cues and textual features to attend to visual regions, achieving full semantic alignment. Residual connections and FFNs stabilize learning, while a subsequent $1{\times}1$ Conv and global MHSA-FFN block further integrate local and global context, yielding robust multimodal representations for grasp reasoning.}
  \label{fig:feature_fusion}
\end{figure*}

\subsection{Dual Cross Vision-language Fusion Bottleneck}
To effectively combine visual and language features for grasp pose prediction, we propose a dual cross vision-language fusion (DCVLF) mechanism that integrates image features with word-level features.
Specifically, the DCVLF integrates visual and textual representations through multi-head attention mechanisms~\cite{vaswani2023attentionneed} for cross-modal fusion. It enables the model to accurately identify and localize target regions in images according to natural language instructions. 
Image features extracted from the visual branch are combined with word-level embeddings. 
The attention mechanism~\cite{vaswani2023attentionneed}, widely adopted in cross-modal fusion, allows models to selectively focus on salient elements within a sequence to capture intrinsic dependencies. 
Given an input sequence \( X \in \mathbb{R}^{n \times d} \) (with sequence length \( n \) and feature dimension \( d \)), attention first projects the sequence into queries \( Q \), keys \( K \), and values \( V \):

\begin{equation}
Q = XW_Q, \quad K = XW_K, \quad V = XW_V.
\end{equation}
where \( W_Q, W_K, W_V \) are learnable projection matrices. Typically, \( d_q = d_k = d_v = d \). The attention operation is then defined as:

\begin{equation}
\text{Attention}(Q, K, V) = \text{SoftMax}\left(\frac{QK^\top}{\sqrt{d}}\right)V.
\end{equation}
where \( QK^\top \) forms an \( n \times n \) similarity matrix indicating inter-element relevance. 
The scaling factor \( 1/\sqrt{d} \) stabilizes gradients, while the row-wise SoftMax converts scores into probability weights. 
The resulting weighted sum of \( V \) emphasizes contextually important elements, allowing explicit modeling of long-range dependencies.

As illustrated in Figure~\ref{fig:feature_fusion}, our DCVLF deeply integrates image and text features. First, image features are enriched with sinusoidal positional encoding and processed through a multi-head self-attention (MHSA) block to capture inter-element relevance representations. Because self-attention is permutation-invariant, positional encoding is essential to preserve geometric structure. Embedding spatial coordinates ensures that attention depends on both content and position, maintaining topological sensitivity. This pre-fusion MHSA modules long-range spatial dependencies, globally reweighting convolutional features and reducing intra-modal redundancy. It also enhances the discriminability of visual queries, ensuring that subsequent cross-modal attention focuses on semantic alignment rather than spatial inference.
For text, we directly use the word-level outputs from the CLIP encoder, which already incorporate positional and contextual encoding. The DCVLF then fuses visual and textual information symmetrically:

\begin{equation}
\tilde{I} = \operatorname{SoftMax}\!\left(\frac{(\hat{I}W_Q)(TW_K)^\top}{\sqrt{d_k}}\right)TW_V,
\end{equation}

\begin{equation}
\tilde{T} = \operatorname{SoftMax}\!\left(\frac{(TW_Q)(\hat{I}W_K)^\top}{\sqrt{d_k}}\right)\hat{I}W_V.
\end{equation}

In the first path, visual features attend to language cues, enriching visual semantics. In the second path, textual features attend to visual regions, enhancing visual grounding. This bidirectional interaction ensures complete multimodal alignment and mitigates information bias inherent in unidirectional fusion.
Each path is followed by residual connections, layer normalization, and an independent feed-forward network (FFN) to enhance nonlinearity and training stability. 
The two enhanced features are concatenated and refined via a \(1\times1\) convolution and ReLU activation:

\begin{equation}
R = \operatorname{ReLU}( \operatorname{Conv_{1\times 1}}(\operatorname{Concat}(\tilde{I},\tilde{T}))).
\end{equation}

Finally, MHSA and FFN blocks are used to improve fine-grained representation learning:

\begin{equation}
H = \operatorname{LN}(R + \operatorname{MHSA}(R)), \quad
O = \operatorname{LN}(H + \operatorname{FFN}(H)).
\end{equation}


\begin{figure*}[t!]
  \centering
  \includegraphics[width=0.95\linewidth]{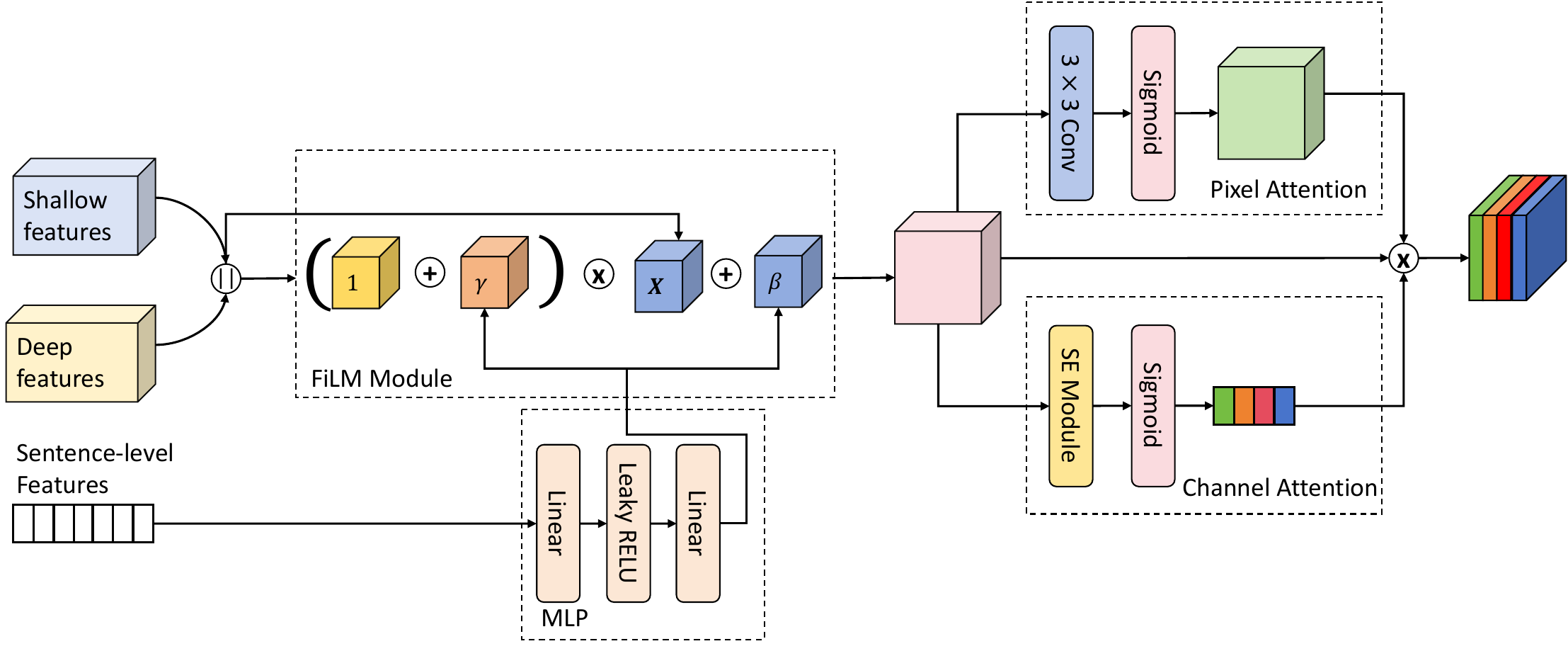}
  \caption{Structure of the proposed LMAFN. The module integrates sentence-level linguistic context into the decoding pathway via FiLM-based feature modulation and dual attention refinement. Low-level spatial features and high-level semantic features are hierarchically fused, while pixel-wise and channel-wise attentions selectively enhance regions relevant to the instruction. This design ensures progressive semantic alignment and accurate reconstruction of fine-grained structural details during up-sampling.}
  \label{fig:lmdafn}
\end{figure*}

\subsection{Hierarchical Language-guided Up-sampling}


The decoding process reconstructs full-resolution outputs from fused multimodal features. We integrate language-aware conditioning and multidimensional attention mechanisms into the up-sampling process. Specifically, Feature-wise Linear Modulation (FiLM) injects linguistic context into the visual domain by adjusting feature activations according to sentence-level semantics. Subsequently, a multidimensional attention module with channel and spatial branches re-weights the feature maps, enhancing regions relevant to the instruction while suppressing background noise. The up-sampling is progressively performed across multiple scales, where language and vision features are fused at each scale. This hierarchical fusion ensures coherent alignment between textual intent and visual details.


\fakeparagraph{Feature-wise Linear Modulation (FiLM).}
FiLM~\cite{perez2018film} introduces lightweight conditioning by dynamically modulating feature channels with linguistic information. The sentence-level text embeddings $s \in \mathbb{R}^{d_{\text{text}}}$ are transformed via a two-layer MLP to generate modulation coefficients:
\begin{equation}
[\gamma, \beta] = W_2\operatorname{LReLU}(W_1 s).
\end{equation}
where $\gamma, \beta \in \mathbb{R}^{C \times 1 \times 1}$ and $\operatorname{LReLU}$ denotes the LeakyReLU activation. The feature maps $x \in \mathbb{R}^{C' \times H' \times W'}$ are modulated as:
\begin{equation}
\operatorname{FiLM}(x, s) = (1 + \gamma) \odot x + \beta.
\end{equation}
where $\odot$ denotes channel-wise multiplication. The additive offset in $(1 + \gamma)$ preserves information when $\gamma=0$, while $\beta$ injects language-dependent biases. Unlike spatial concatenation methods, FiLM maintains the tensor structure, providing efficient channel-wise semantic gating that aligns global sentence semantics with visual activations.

As shown in Figure~\ref{fig:lmdafn}, shallow features $x_s \in \mathbb{R}^{C' \times H' \times W'}$ preserve spatial structures, while deep features $x_d \in \mathbb{R}^{C' \times H' \times W'}$ encode high-level semantics. They are concatenated:
\begin{equation}
x_{\text{cat}} = \operatorname{Concat}(x_d, x_s) \in \mathbb{R}^{2C' \times H' \times W'}.
\end{equation}

Then, FiLM, conditioned by sentence-level embeddings, refines $x_{\text{cat}}$ to emphasize task-relevant semantics. 
\begin{equation}
x_{\text{FiLM}} = \operatorname{FiLM}(x_{\text{cat}}, s).
\end{equation}

\fakeparagraph{Language-guided Multidimensional Attention Fusion Network (LMAFN).}
Our LMAFN extends channel-wise recalibration~\cite{hu2018squeeze} with pixel-level and linguistic guidance. Specifically, a 3$\times$3 convolution followed by sigmoid activation generates pixel-level attention maps:
\begin{equation}
\mathbf{A}_{\text{pixel}} = \sigma(\text{Conv}_{3 \times 3}(x_{\text{FiLM}})) \in \mathbb{R}^{C \times H \times W}.
\end{equation}

Parallel to pixel-level attention, the FiLM-enhanced feature $x_{\text{FiLM}}$ is first processed by global average pooling to obtain $z$. Then it passed through a shared two-layer MLP:
\begin{equation}
z = \mathrm{GAP}(x_{\mathrm{FiLM}}),
\end{equation}
\begin{equation}
A_{\mathrm{channel}} = \sigma\!\left( W_2 \, \delta\!\left( W_1 z \right) \right).
\end{equation}
where $W_1$ and $W_2$ denote MLP weights. Weight sharing is used to ensure consistent semantic transformations for both modalities.
Finally, the outputs integrate both attention maps:
\begin{equation}
x_{\text{out}} = x_{\text{FiLM}} \odot A_{\text{pixel}} \odot A_{\text{channel}}.
\end{equation}
\begin{figure*}[t!]
  \centering
  \includegraphics[width=0.8\linewidth]{Images/lgdc.pdf}
  \caption{Structure of the Language-conditioned Dynamic Convolution Head(LDCH). The module adopts a kernel-mixture formulation in which sentence-level features guide the weighted combination of multiple convolution experts to generate language-conditioned coarse predictions.
  }
  \label{fig:lgdc}
\end{figure*}

\subsection{Coarse Mask and Grasp Prediction Head}

To specialize output predictions according to textual instructions, we introduce a language-guided dynamic convolution module for the coarse mask and grasp prediction heads. The key idea is to use sentence-level features to predict a mixture over a small set of learnable convolution experts~\cite{chen2020dynamic}, and to form Language-conditioned kernels via a weighted combination, as illustrated in the Figure~\ref{fig:lgdc}.

\fakeparagraph{Task-wise feature preparation.}
Given the feature map $x_{\text{out}}$, we first expand its channel dimension using a $1{\times}1$ convolution:
\begin{equation}
\mathbf{F} = \text{Conv}_{1\times1}(x_{\text{out}}) \in \mathbb{R}^{B \times 5C \times H \times W}.
\end{equation}
The output is then partitioned into five task-specific feature groups through a fixed channel-wise slicing operation:
\begin{equation}
\mathbf{F} = [F_{\text{mask}}; F_{\text{quality}}; F_{\text{sin}}; F_{\text{cos}}; F_{\text{width}}],
\quad F_t \in \mathbb{R}^{B \times C \times H \times W},
\end{equation}
corresponding to the segmentation mask, grasp quality, grasp angle (sine and cosine), and gripper width prediction heads. For efficient per-sample dynamic convolution, each $F_t$ is reshaped for grouped
convolution:
\begin{equation}
\tilde{F}_t = \text{reshape}(F_t) \in \mathbb{R}^{1 \times (B \cdot C) \times H \times W}.
\end{equation}

\fakeparagraph{Language-guided kernel mixing.}
For each sample, we extract a sentence-level embedding $s_i \in \mathbb{R}^{d}$ and predict a distribution over $K$ kernel experts using an MLP followed by a softmax:
\begin{equation}
\pi_i = \text{Softmax}(\text{MLP}(s_i)), \quad \pi_i \in \mathbb{R}^{K},
\end{equation}
where $\text{MLP}(\cdot)$ is a two-layer perceptron with a LeakyReLU nonlinearity. For each prediction head $t$, we maintain a bank of $K$ learnable convolution experts:
\begin{equation}
\{W_{t}^{(k)}, b_{t}^{(k)}\}_{k=1}^{K}, \quad
W_{t}^{(k)} \in \mathbb{R}^{C \times 3 \times 3}, \;\; b_{t}^{(k)} \in \mathbb{R}.
\end{equation}
Given the mixture weights $\pi_i$, the language-conditioned kernel and bias are formed via a weighted sum:
\begin{equation}
W_{i,t} = \sum_{k=1}^{K} \pi_{i,k} W_{t}^{(k)}, \qquad
b_{i,t} = \sum_{k=1}^{K} \pi_{i,k} b_{t}^{(k)}.
\end{equation}
Stacking all instances yields the dynamic kernel and bias bank:
\begin{equation}
\begin{aligned}
W_{t,\text{d}} &= \{W_{1,t}, \dots, W_{B,t}\} \in \mathbb{R}^{B \times C \times 3 \times 3}, \\
b_{t,\text{d}} &= [b_{1,t}, \dots, b_{B,t}] \in \mathbb{R}^{B}.
\end{aligned}
\end{equation}

\fakeparagraph{Coarse prediction via grouped convolution.}
We apply the instruction-conditioned parameters through a grouped convolution with the number of groups equal to the batch size:
\begin{equation}
Y_t = \text{GroupConv2D}(\tilde{F}_t, W_{t,\text{d}}, b_{t,\text{d}}, \text{groups}=B),
\end{equation}
and reshape the output back to the standard batch format
$Y_t \in \mathbb{R}^{B \times 1 \times H \times W}$.
The same mechanism is used for all five heads, producing language-conditioned coarse predictions for the mask and grasp parameter maps.


\begin{figure}[t!]
  \centering
  \includegraphics[width=\linewidth]{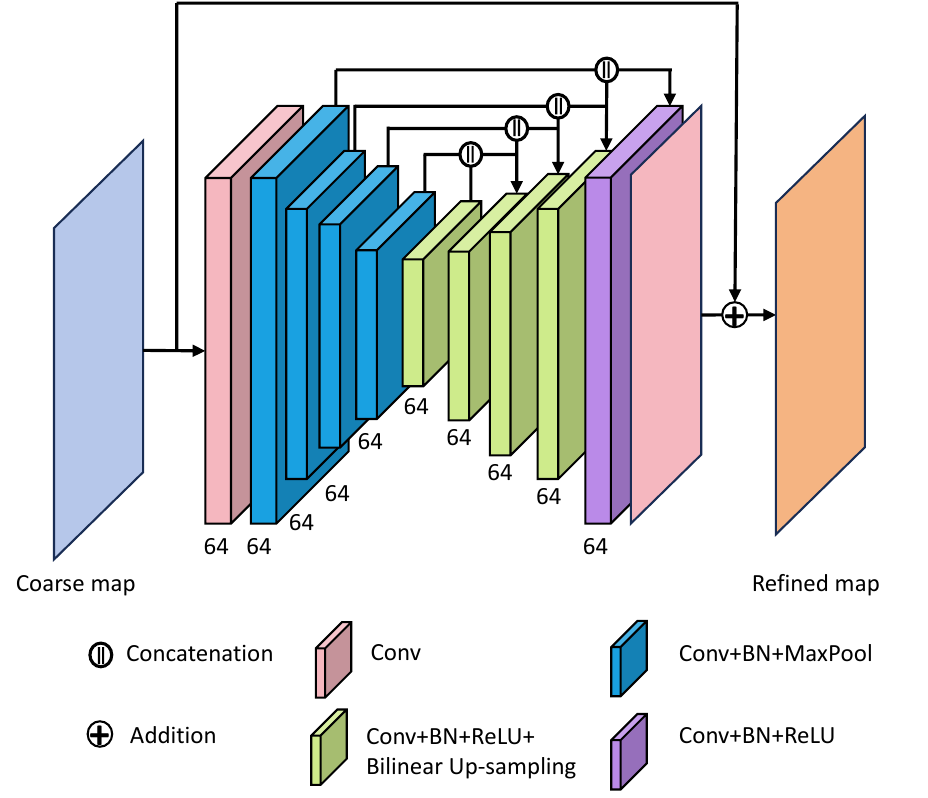}
  \caption{Structure of the Refinement Module (RM). The RM adopts a lightweight U-Net-style structure to learn residual corrections over coarse predictions.
  }
  \label{fig:rrm}
\end{figure}

\subsection{Mask and Grasp Refinement Modules}

We employ a lightweight U-Net-style architecture, which learns residual corrections to coarse predictions rather than predicting the final outputs directly. This residual learning design stabilizes training and preserves the information from the preceding prediction stage, as illustrated in Figure~\ref{fig:rrm}.
Specifically, we employ two refinement branches: the mask refinement module refines the segmentation mask, and the grasp refinement module jointly refines grasp quality, angle (sine and cosine), and gripper width. 
Given coarse predictions,
\begin{equation}
Y^{\text{coarse}}_{\text{mask}} \in \mathbb{R}^{B \times 1 \times H \times W}, \quad
Y^{\text{coarse}}_{\text{grasp}} \in \mathbb{R}^{B \times 4 \times H \times W},
\end{equation}
each branch learns a residual correction $\Delta$ that refines the coarse predictions:
\begin{align}
Y_{\text{mask}} &= Y^{\text{coarse}}_{\text{mask}} + \text{Mask}_{\text{refine}}(Y^{\text{coarse}}_{\text{mask}}), \\
Y_{\text{grasp}} &= Y^{\text{coarse}}_{\text{grasp}} + \text{Grasp}_{\text{refine}}(Y^{\text{coarse}}_{\text{grasp}}).
\end{align}

As shown in Figure~\ref{fig:rrm}, each refinement module follows a symmetric encoder-decoder structure. The encoder consists of four convolutional blocks with max-pooling for context aggregation, while the decoder entails mirrored convolutional blocks with bilinear up-sampling and skip connections to recover spatial details. The outputs are one-channel (mask) or four-channel (grasp) residual maps aligned with the input resolution. All refinement operations are applied on logits before activation, ensuring gradient flow through both coarse and refined prediction paths for stable end-to-end optimization. This coarse-to-fine framework, coarse estimation followed by residual refinement, improves precision and semantic coherence, especially in cases involving occluded or irregularly shaped objects, producing sharper masks and more stable grasp orientation predictions.

\subsection{Loss Function}
To facilitate the joint optimization of language-guided segmentation and grasp prediction, we design a comprehensive multi-stage, multi-task loss. The objective supervises both the coarse prediction branch and the refinement branch, and couples binary mask segmentation with continuous grasp regression into a unified training framework.

\fakeparagraph{Multi-stage Deep Supervision.}
Our architecture employs a refinement module with residual connections to progressively enhance the grasp predictions. To fully utilize the learning capacity of intermediate feature stages and stabilize optimization, we adopt deep supervision on both the coarse and refined outputs. The overall training objective is formulated as:
\begin{equation}
\mathcal{L}_{\text{total}} = \mathcal{L}_{\text{refine}} + \lambda \cdot \mathcal{L}_{\text{coarse}}.
\end{equation}
where $\mathcal{L}_{\text{refine}}$ denotes the loss imposed on the final refined predictions, and $\mathcal{L}_{\text{coarse}}$ is an auxiliary loss applied to the coarse predictions. The coefficient $\lambda$ controls the contribution of the coarse supervision. In practice, we set $\lambda = 0.5$ to provide sufficient gradient signals to the coarse stage while preventing it from dominating the optimization of the refinement branch.
Each stage-specific loss decomposes into a segmentation term and a grasp regression term:
\begin{equation}
\mathcal{L}_{\text{refine}} = \mathcal{L}_{\text{mask}}^{\text{refine}} + \mathcal{L}_{\text{grasp}}^{\text{refine}},
\end{equation}
\begin{equation}
\mathcal{L}_{\text{coarse}} =  \mathcal{L}_{\text{mask}}^{\text{coarse}} + \mathcal{L}_{\text{grasp}}^{\text{coarse}}.
\end{equation}
where $\mathcal{L}_{\text{mask}}^{\text{refine}}$ and $\mathcal{L}_{\text{mask}}^{\text{coarse}}$ supervise the binary segmentation masks at refined and coarse resolutions, respectively, and $\mathcal{L}_{\text{grasp}}^{\text{refine}}$ and $\mathcal{L}_{\text{grasp}}^{\text{coarse}}$ supervise the corresponding dense grasp predictions. This multi-stage deep supervision encourages the network to produce semantically meaningful, language-consistent predictions at early stages, while the refinement module focuses on correcting residual errors and improving prediction accuracy.

\fakeparagraph{Segmentation Mask Loss.}
The segmentation branch predicts a binary mask indicating the language-referred object or region to be grasped. Let $Y_{\text{mask}} \in \mathbb{R}^{H \times W}$ denote the predicted logit map and $M \in \{0,1\}^{H \times W}$ the ground-truth binary mask, where $M_{uv} = 1$ indicates that pixel $(u,v)$ belongs to the target region specified by the language query. We use a weighted binary cross-entropy loss with logits:
\begin{equation}
\begin{aligned}
\mathcal{L}_{\text{mask}} 
&= -\frac{1}{\sum_{u,v} w_{uv}}
\sum_{u=1}^{H} \sum_{v=1}^{W} 
w_{uv} \big[
    M_{uv} \log \sigma(Y_{uv}) \\
&\quad + (1 - M_{uv}) \log (1 - \sigma(Y_{uv}))
\big].
\end{aligned}
\end{equation}
where $\sigma(\cdot)$ is the sigmoid activation that maps logits to probabilities. The normalization by $\sum_{u,v} w_{uv}$ keeps the loss scale consistent across images with different spatial sizes and varying object extents.
To alleviate class imbalance between foreground and background pixels, which is particularly severe when the language-referred object occupies only a small portion of the image, we introduce a simple yet effective foreground reweighting scheme:
\begin{equation}
w_{uv} = \beta \cdot M_{uv} + 1.
\end{equation}

In this formulation, background pixels receive a weight of $1$, while foreground pixels are weighted by $1 + \beta = 1.5$. This increases the contribution of correctly segmenting the language-targeted region and prevents the model from converging to trivial background-dominant solutions. The hyperparameter $\beta$ can be tuned according to the foreground-to-background ratio of the dataset; higher values impose stronger penalties on misclassified foreground pixels.
Note that the segmentation mask serves a dual purpose: it not only evaluates how well the model segments the language-specified object, but also defines the set of valid pixels used for supervising grasp parameters, thereby tightly coupling spatial localization and grasp prediction.

\fakeparagraph{Grasp Regression Loss.}
The grasp head predicts grasp parameters for each spatial location, conditioned jointly on visual features and language-guided fusion features. While a planar grasp can be conceptually defined as 
$g=\{(x,y),\theta,w,q\}$,
in our implementation we adopt a learning-friendly parameterization:
\begin{equation}
G = \{ q, \cos(2\theta), \sin(2\theta), w \} \in \mathbb{R}^4.
\end{equation}
where $q \in [0,1]$ denotes the grasp quality score, $\theta \in [-\pi/2, \pi/2]$ is the in-plane grasp angle, and $w > 0$ represents the gripper width required to execute the grasp. The angle is encoded by its double-angle trigonometric form $(\cos(2\theta), \sin(2\theta))$ to avoid discontinuities at $\theta = \pm \pi/2$ and to ensure smooth optimization over angular space.
Let $\hat{G} \in \mathbb{R}^{H \times W \times 4}$ and $G \in \mathbb{R}^{H \times W \times 4}$ denote the predicted and ground-truth grasp tensors, respectively. We compute the grasp regression loss only over valid grasp locations, which are defined by the ground-truth mask:
\begin{equation}
\mathcal{V} = \{ i \mid M_i = 1 \}.
\end{equation}
where $i$ indexes spatial locations in the image. The loss is given by:
\begin{equation}
\mathcal{L}_{\text{grasp}} = \frac{1}{|\mathcal{V}|} 
\sum_{i \in \mathcal{V}} 
\sum_{m \in \mathcal{M}} 
 \text{SmoothL1}\big(\hat{G}_i^m - G_i^m\big).
\end{equation}
where $\mathcal{M} = \{ q, \cos(2\theta), \sin(2\theta), w \}$ indexes the grasp components. This masking strategy ensures that supervision focuses on language-relevant object regions and avoids penalizing predictions on irrelevant background areas.
We use the Smooth L1 loss due to its robustness to outliers and stable gradients:
\begin{equation}
\text{SmoothL1}(x) = 
\begin{cases} 
\frac{1}{2}(\sigma x)^2, & \text{if } |x| < \frac{1}{\sigma^2} \\
|x| - \frac{1}{2\sigma^2}, & \text{otherwise} .
\end{cases}
\end{equation}
where $\sigma$ controls the transition point between the quadratic and linear regimes. We set $\sigma = 1$, which results in a quadratic penalty for small errors ($|x| < 1$), encouraging precise grasp predictions, and a linear penalty for larger deviations, limiting the influence of noisy or ambiguous annotations. This is particularly important in language-guided grasping, where fine-grained object parts (e.g., “handle” vs. “body”) may be annotated with some degree of uncertainty.

Overall, the proposed loss formulation tightly couples segmentation and grasp prediction under language guidance: the mask loss drives the network to localize the language-referred object, while the regression loss encourages accurate and semantically consistent grasp parameters within that region. The multi-stage supervision further ensures that both coarse and refined predictions benefit from strong training signals, leading to robust and stable convergence.

\section{Experimental Results}

\subsection{Dataset}
To evaluate our model's performance, we use the OCID-VLG~\cite{tziafas2023languageguidedrobotgraspingclipbased} and Grasp-Anything++~\cite{vuong2024language} datasets, which are the most comprehensive benchmarks available for language-guided grasp detection.

\fakeparagraph{OCID-VLG.} 
OCID-VLG is constructed on top of OCID~\cite{8793917,ainetter2021end}, extending it to support language-driven grasping tasks. It is derived from 1,763 unique OCID scenes and comprises nearly 90k fully annotated tuples. Each tuple includes an RGB image, a natural language instruction specifying the target object, a pixel-level segmentation mask of the target, and a set of valid grasp rectangles. To assess our model’s accuracy, generalization, and zero-shot capabilities, we conduct experiments using the following three dataset splits:
\begin{itemize}
  \item \textbf{Multiple-Split.} This is a standard random scene split in which both the object classes and instances appearing in the test set are also present during training. We use this split to evaluate the model’s performance under conventional conditions.

  \item \textbf{Novel-Instances Split.} In this split, the test set includes object instances that are unseen during training, although their corresponding classes are observed. This setup is used to evaluate the model’s instance-level generalization-its ability to transfer knowledge from seen instances to novel ones within the same class.

  \item \textbf{Novel-Classes Split.} This is the most challenging split, where the test set contains object classes that never appear during training. We use this split to evaluate the model’s zero-shot capability-its ability to reason about and localize novel classes based solely on language descriptions and prior visual knowledge, without having seen any corresponding examples.
\end{itemize}

\fakeparagraph{Grasp-Anything++.} 
We further train and evaluate our model on the synthetic Grasp-Anything++ dataset~\cite{vuong2024language}, an extension of the original Grasp-Anything dataset~\cite{vuong2023graspanything}. Grasp-Anything is generated using foundation models and contains 1 million samples featuring over 3 million diverse objects, along with 10 million grasping instructions paired with corresponding ground truth. Experiments on Grasp-Anything++ enable us to assess whether the proposed architecture can perform effectively in large-scale synthetic environments with procedurally generated scenes and full-sentence prompts.

\subsection{Evaluation Metrics}

To evaluate our model's performance across different tasks, we adopt a set of established metrics that measure its effectiveness in two domains: referring object segmentation and referring grasp detection.

\fakeparagraph{Grasp Detection Metrics.}
For referring grasp detection, where the goal is to identify valid grasp candidates in an image, we adopt the rectangle-based evaluation protocol used in prior works~\cite{kumra2020antipodal, cao2021residual,tziafas2023languageguidedrobotgraspingclipbased}. A predicted grasp is considered correct if it satisfies both of the following conditions:

\begin{itemize}
    \item \textbf{Angle Difference:} The absolute difference between the predicted grasp rectangle’s orientation and that of the ground-truth rectangle is less than $30^\circ$.
    
    \item \textbf{Jaccard Index:} The Intersection over Union (IoU) between the predicted grasp rectangle $g_p$ and a ground-truth rectangle $g_t$ exceeds 0.25. The IoU, also known as the Jaccard index, is defined as
    \begin{equation}
        J(g_p, g_t) = \frac{|g_p \cap g_t|}{|g_p \cup g_t|} > 0.25 .
    \end{equation}
\end{itemize}
To evaluate the model’s ability to predict a valid grasp corresponding to a given referring expression, we report the $J@N$ metric. Following the protocol in~\cite{tziafas2023languageguidedrobotgraspingclipbased}, a prediction is considered successful if the top-N ranked predicted grasp:

\begin{itemize}
    \item Refers to the correct object as specified by the language instruction.
    \item Satisfies both the angle criterion (orientation difference less than $30^\circ$) and the IoU (Jaccard index) criterion (greater than 0.25) with respect to the ground-truth grasp on the referred object.
\end{itemize}

These evaluations capture the model’s ability to integrate visual grounding with grasp estimation in language-conditioned tasks.

\fakeparagraph{Referring Segmentation Metrics.}
To assess the model’s ability to segment the object specified by a textual instruction, we follow the evaluation protocol in~\cite{wang2022cris, tziafas2023languageguidedrobotgraspingclipbased}. Two metrics are employed:

\begin{itemize}
    \item \textbf{Intersection over Union(IoU):} The average IoU between the predicted and ground-truth segmentation masks.
    
    \item \textbf{Precision@X (Pr@X):} The percentage of test instances for which the IoU exceeds a threshold $X$. Results are reported for $X \in \{0.5, 0.6, 0.7, 0.8, 0.9\}$.
\end{itemize}

\subsection{Training Setup}

Our experiments are implemented in PyTorch 2.4.0~\cite{paszke2019pytorch} with CUDA 12.4 support. All models are trained end-to-end on a single NVIDIA H100 GPU with 80 GB memory. We use the AdamW optimizer~\cite{loshchilov2017decoupled} with a base learning rate of $1\times10^{-4}$ and a weight decay of 0.06. The learning rate follows a cosine annealing schedule~\cite{loshchilov2016sgdr} with a linear warm-up during the first 2,000 iterations~\cite{he2016deep}, starting from $1\times10^{-5}$ to improve early-stage training stability. Models are trained for 50 epochs with a batch size of 72.


\subsection{Results on OCID-VLG}



\begin{table*}[t!]
\centering
\caption{Evaluation results of different methods on the OCID-VLG dataset with the Multi-Split setting. Bold and underlined numbers denote the best and second-best results, respectively.}
\begin{tabular}{l|cc|ccccccc}
\toprule
\textbf{Model} & \textbf{J@1} & \textbf{J@5} & \textbf{IoU} & \textbf{Pr@50}& \textbf{Pr@60} & \textbf{Pr@70} & \textbf{Pr@80} & \textbf{Pr@90} \\
\midrule
GT-Grounding & 28.7 & - & - & - & - & - & - & - \\
GT-Masks + CLIP~\cite{radford2021learningtransferablevisualmodels} & 11.9 & - & 35.0 & 35.0 & 35.0 & 35.0 & 35.0 & 35.0 \\
GR-ConvNet~\cite{kumra2020antipodal} + CLIP~\cite{radford2021learningtransferablevisualmodels} & 9.7 & 15.4 & 31.3 & 21.0 & 11.6 & 5.5 & 2.4 & 0.5 \\
SAM~\cite{kirillov2023segment} + CLIP~\cite{radford2021learningtransferablevisualmodels}  & 7.2 & - & 25.7 & 29.3 & 28.5 & 27.4 & 22.7 & 9.1 \\
GLIP~\cite{li2022grounded} + SAM~\cite{kirillov2023segment}  & 10.7 & -  & 30.3 & 34.7 & 34.1 & 33.5 & 28.6 & 11.7 \\
Det-Seg~\cite{ainetter2021end} + CLIP~\cite{radford2021learningtransferablevisualmodels} & 28.1 & -  & 29.0 & 27.2 & 20.9 & 17.5 & 17.2 & 16.0 \\
SSG~\cite{xu2023instance} + CLIP~\cite{radford2021learningtransferablevisualmodels} & 33.5 & -  & 33.6 & 35.6 & 35.6 & 35.5 & 35.5 & \textbf{32.8} \\
CLIP~\cite{radford2021learningtransferablevisualmodels} + MHSA~\cite{vaswani2023attentionneed} & 82.5 & 87.7 & 71.3 & 93.2 & 84.4 & 54.5 & 27.7 & 0.1 \\
EfficientGrasp~\cite{cao2022efficient}  + CLIP~\cite{radford2021learningtransferablevisualmodels} & 79.2 & 82.4 & 80.2 & 87.4 & 71.8 & 43.1 & 32.1 & 15.6 \\
CROG~\cite{tziafas2023languageguidedrobotgraspingclipbased}   & 77.2 & - & 81.1 & 96.9 & \textbf{94.8} & 87.2 & 64.1 & 16.4 \\
\midrule
     Ours  & \textbf{85.4} & \textbf{90.2} & \textbf{83.1} & \textbf{98.0} & \underline{93.7} & \textbf{91.3} &\textbf{66.7} & \underline{26.1} \\
\bottomrule
\end{tabular}
\label{tab:ocid-vlg-results-multi}
\end{table*}

\fakeparagraph{Multi-Split.}
The evaluation on the OCID-VLG dataset under the Multi-Split setting provides a comprehensive benchmark for model performance, simulating a scenario in which both object classes and instances in the test set are seen during training. As shown in Table~\ref{tab:ocid-vlg-results-multi}, our proposed method outperforms the recent method, CROG~\cite{tziafas2023languageguidedrobotgraspingclipbased}, achieving an IoU of \textbf{83.1\%} for segmentation and a J@1 of \textbf{85.4\%} for top-1 grasp prediction, compared to CROG's 81.1\% IoU and 77.2\% J@1. These results validate the effectiveness of our approach.



\begin{table}[t!]
\centering
\caption{Evaluation results of different methods on the OCID-VLG dataset with Novel-Instances Split setting.}
\label{tab:ocid-vlg-novel}
\begin{tabular}{l|cc|cccc}
\toprule
\textbf{Model}  & \textbf{J@1} & \textbf{J@5} & \textbf{IoU} & \textbf{Pr@50} & \textbf{Pr@70} & \textbf{Pr@90} \\
\midrule
CROG~\cite{tziafas2023languageguidedrobotgraspingclipbased}  & 48.3 & 52.0 & 61.3 & 74.8 & 52.2 & 1.7 \\
Ours  & \textbf{57.6} & \textbf{64.1} & \textbf{67.6} & \textbf{76.4} & \textbf{61.9} & \textbf{17.1} \\
\bottomrule
\end{tabular}
\end{table}

\fakeparagraph{Novel-Instances Split.}
To assess the models’ ability to generalize to new instances of object classes seen during training, we adopt the Novel-Instances Split. Table~\ref{tab:ocid-vlg-novel} reports the performance of each model under this setting. The recent method, CROG~\cite{tziafas2023languageguidedrobotgraspingclipbased}, exhibits a substantial performance decline, whereas our model demonstrates stronger generalization capabilities. This degradation in CROG can be attributed to its strategy of fine-tuning the entire CLIP~\cite{radford2021learningtransferablevisualmodels} backbone on a task-specific dataset rather than freezing it. While such full fine-tuning may improve in-domain performance, it increases the risk of overfitting and can induce catastrophic forgetting of the broad visual-semantic knowledge acquired during large-scale pre-training. As a result, CROG struggles when encountering novel instances, leading to the observed performance drop.


\begin{table}[t!]
\centering
\caption{Evaluation results of different methods on the OCID-VLG dataset with Novel-Classes Split setting.}
\label{tab:ocid-vlg-novelclass}
\begin{tabular}{l|cc|cccc}
\toprule
\textbf{Model}  & \textbf{J@1} & \textbf{J@5} & \textbf{IoU} & \textbf{Pr@50} & \textbf{Pr@70} & \textbf{Pr@90} \\
\midrule
CROG~\cite{tziafas2023languageguidedrobotgraspingclipbased}  & 14.6 & 15.2 & 40.4 & 46.5 & 23.9 & 0.1 \\
Ours    & \textbf{46.0} & \textbf{54.4} & \textbf{63.1} & \textbf{72.5} & \textbf{54.8} & \textbf{19.4} \\
\bottomrule
\end{tabular}
\end{table}

\fakeparagraph{Novel-Classes Split.}
The Novel-Classes Split provides a rigorous evaluation of a model's language understanding and reasoning capabilities, as it requires handling object classes that are entirely unseen during training. The experimental results are presented in Table~\ref{tab:ocid-vlg-novelclass}. CROG exhibits a substantial decline in performance under this setting, with its J@1 and IoU dropping sharply. As discussed earlier, fully fine-tuning the CLIP~\cite{radford2021learningtransferablevisualmodels} backbone tends to “contaminate” the open-world knowledge acquired during pre-training with task-specific biases, thereby severely weakening the model’s zero-shot generalization ability. In contrast, our proposed method achieves a J@1 of 46.0\% and an IoU of 63.1\%, demonstrating strong robustness when confronted with novel object classes.


\begin{table}[t!]
\centering
\caption{Evaluation results of different methods on the Grasp-Anything++ dataset (full prompt sentence).}
\begin{tabular}{l|cc}
\toprule
\textbf{Model} & \textbf{Seen} & \textbf{Unseen} \\
\midrule
GR-ConvNet~\cite{kumra2020antipodal} + CLIP~\cite{radford2021learningtransferablevisualmodels}  & 0.21 & 0.12 \\
GGCNN~\cite{morrison2018closinglooproboticgrasping} + CLIP~\cite{radford2021learningtransferablevisualmodels} & 0.11 & 0.07 \\
EfficientGrasp~\cite{cao2022efficient} + CLIP~\cite{radford2021learningtransferablevisualmodels}     & 0.33 & 0.18 \\
CLIPort~\cite{shridhar2022cliport}              & 0.29 & 0.19 \\
LGD~\cite{vuong2024language}     & 0.41 & --   \\
\midrule
Ours & \textbf{0.59} & \textbf{0.31} \\
\bottomrule
\end{tabular}
\label{tab:grasp-anything}
\end{table}

\subsection{Results on Grasp-Anything++}

To assess whether our approach is effective on synthetic data, we further train and evaluate our model on the Grasp-Anything++ dataset using full-sentence prompts. This dataset features procedurally generated tabletop scenes with diverse object configurations and natural language instructions, and it provides separate splits for scenes containing seen and unseen objects. The quantitative results are summarized in Table~\ref{tab:grasp-anything}. Our method outperforms prior approaches on both splits. Compared to the second-best method, LGD~\cite{vuong2024language}, our performance improves to 0.59 and 0.31 on the seen and unseen splits, respectively. These results demonstrate that our language-guided grasping framework generalizes well not only to real-world datasets but also to large-scale synthetic environments with full-sentence prompts.

\subsection{Qualitative Evaluations}
\begin{figure}[t!]      
  \centering
  \includegraphics[width=\linewidth]{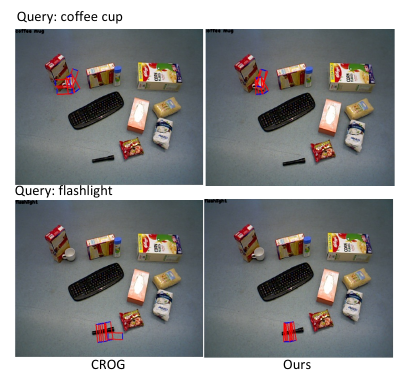}
  \caption{Results of reasoning grasping with key words.}
  \label{fig:key-word}
\end{figure}

\begin{figure}[t!]      
  \centering
  \includegraphics[width=\linewidth]{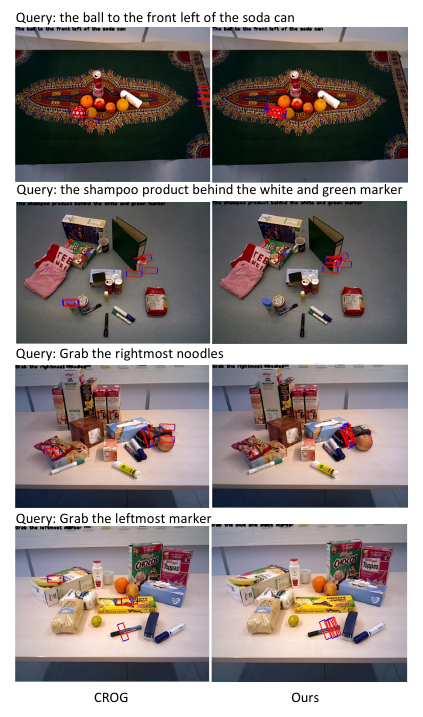}
  \caption{Results of reasoning grasping with location instructions.}
  \label{fig:location}
\end{figure}

\begin{figure}[t!]      
  \centering
  \includegraphics[width=\linewidth]{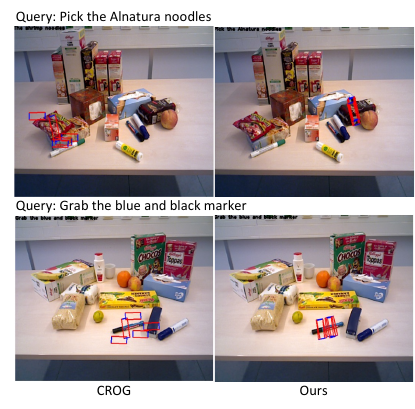}
  \caption{Results of reasoning grasping with direct descriptions.}
  \label{fig:description}
\end{figure}

\begin{figure}[t!]        
  \centering
  \includegraphics[width=1.0\linewidth]{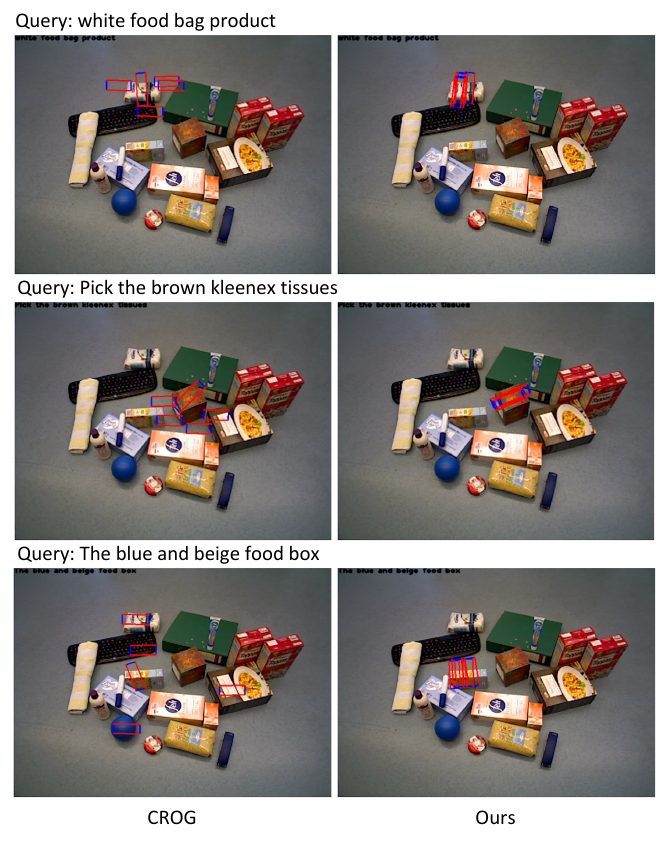}
  \caption{Results of reasoning grasping on the novel instances.}
  \label{fig:five-instances}
\end{figure}

The complexity of natural language instructions also affects model performance. To analyze this effect, we compare how different models behave under varying instruction types, including reasoning grasping with key words, reasoning grasping with location instructions, reasoning grasping with direct descriptions, and reasoning grasping with novel instances. The visualizations of the CROG and our model under three different instruction types are shown in Figure~\ref{fig:key-word}, Figure~\ref{fig:location}, Figure~\ref{fig:description}, and Figure~\ref{fig:five-instances}.

\fakeparagraph{Reasoning Grasping with Key Words.}
For simple instructions containing only object keywords, all models successfully localize the target and generate correct grasp poses, as shown in Figure~\ref{fig:key-word}. This demonstrates that both CROG and our model exhibit basic vision-language alignment when the semantic cues are straightforward.

\fakeparagraph{Reasoning Grasping with Location Instructions.}
When the instruction includes spatial location descriptions (as shown in Figure~\ref{fig:location}), CROG's performance begins to deteriorate, often misidentifying multiple objects as potential grasp targets. In contrast, our model performs reliably across different positional terms. Even when the instruction requires more complex spatial reasoning, our approach consistently interprets and resolves composite spatial relationships correctly.

\fakeparagraph{Reasoning Grasping with Direct Descriptions.}
For instructions that require understanding specific attributes or brand names, our model also outperforms CROG, as illustrated in Figure~\ref{fig:description}.

\fakeparagraph{Reasoning Grasping on the Novel Instances.}
The proposed model further demonstrates strong zero-shot understanding by correctly identifying objects described using non-obvious brand names. The visualizations in Figure~\ref{fig:five-instances} highlight this difference in generalization ability. For the instruction \textit{"Pick the brown kleenex tissues"}, the target refers to a specific tissue box variant not seen during training. CROG fails completely in this scenario, incorrectly localizing its grasp predictions on unrelated objects-an indication of its limited generalization. In contrast, our model is able to correctly identify and approximately localize the intended tissue box.

\begin{figure*}[t!]         
\centering
  \includegraphics[width=0.95\linewidth]{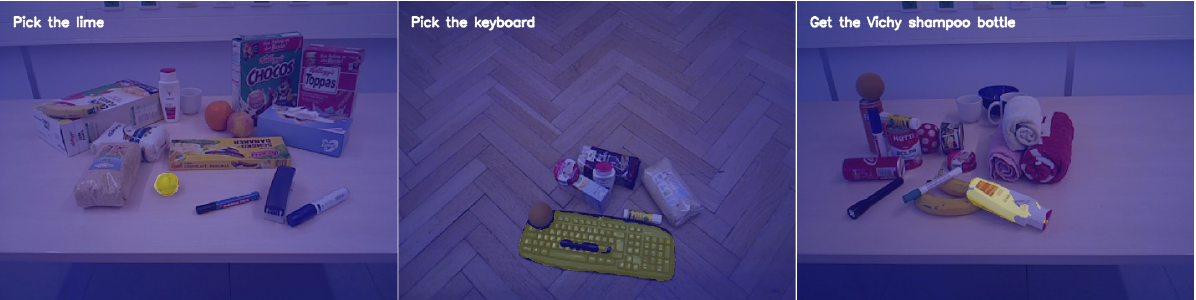}
  \caption{Attention map visualization of our model.}
  \label{fig:attention_com}
\end{figure*}
\fakeparagraph{Attention Map Visualization.}
To further illustrate how our model leverages language to guide visual reasoning, we visualize its attention maps for several representative instruction–scene pairs, as shown in Figure~\ref{fig:attention_com}. The visualizations show that the model consistently attends to task-relevant regions aligned with the textual instructions. For the command “\textit{``Pick the lime"} the attention map focuses precisely on the small lime despite numerous distractors. When instructed to \textit{``Pick the keyboard"}, the attention spreads along the keyboard’s elongated structure, accurately covering its graspable area. For the more fine-grained instruction \textit{``Get the Vichy shampoo bottle"}, the model successfully isolates the target bottle from several visually similar containers.


\begin{table*}[t!]
\centering
\caption{Comparison of model complexity and efficiency.}
\begin{tabular}{l|ccccc}
\toprule
\textbf{Model} & \textbf{Parameters (M)} & \textbf{FLOPs (G)} & \textbf{MACs (G)} & \textbf{Inference Time (ms)} & \textbf{Trainable Parameters (M)} \\
\midrule
GR-ConvNet~\cite{kumra2020antipodal} + MHSA~\cite{vaswani2023attentionneed}  & 55.77 & 111.53  & 64.81   & 21.00  & 52.21 \\
CROG~\cite{tziafas2023languageguidedrobotgraspingclipbased}   & 147.11  & 223.50  & 111.72   & 39.11  & 87.11 \\
\midrule
Ours  & 171.76  & 289.21  & 144.61  & 43.00  & 67.70 \\
\bottomrule
\end{tabular}
\label{tab:model-complexity}
\end{table*}

\subsection{Efficiency Analysis}

In robotic applications, efficiency is as important as accuracy. Table~\ref{tab:model-complexity} summarizes the computational costs of the evaluated models, including parameter size, FLOPs, MACs, and single-sample inference time. When considered alongside the accuracy results in Table~\ref{tab:ocid-vlg-results-multi}, these metrics highlight the trade-off between model performance and computational cost. The results demonstrate that our proposed method strikes an effective balance between efficiency and accuracy, achieving an inference time of 43.00 ms with a parameter budget of 171.76 M, while delivering superior grasping performance.

\begin{table}[t!]
\centering
\caption{Ablation study on the key components.}
\label{tab:ablation-lgrd}

\begin{tabular}{c c c c | c c}
\toprule
\textbf{DCVLF} & \textbf{LMAFN} & \textbf{RM} & \textbf{LDCH} & \textbf{IoU} & \textbf{J@1} \\
\midrule
 &  &  &  & 31.30 & 9.73 \\
\checkmark &  &  &  & 71.34 & 79.36 \\
\checkmark & \checkmark &  &  & 80.16 & 82.27 \\
\checkmark & \checkmark & \checkmark &  & 82.29 & 84.53 \\
\checkmark & \checkmark & \checkmark & \checkmark & \textbf{83.14} & \textbf{85.36} \\
\bottomrule
\end{tabular}
\end{table}

\subsection{Ablation Study}

Table~\ref{tab:ablation-lgrd} reports a step-by-step ablation where we progressively add each key component to the baseline. The baseline achieves only 31.30\% IoU and 9.73\% J@1, suggesting that simple feature concatenation fails to capture the fine-grained alignment required by language-guided grasping. After introducing Dual Cross Vision-language Fusion Bottleneck (DCVLF), performance increases to 71.34\% IoU and 79.36\% J@1 (+40.04/+69.63), indicating that DCVLF is effective establishing vision-language fusion.
Building on this, adding Hierarchical Language-guided Up-sampling with Language-guided Multidimensional Attention Fusion Network (LMAFN) further improves results to 80.16\% IoU and 82.27\% J@1 (+2.57/+0.85), demonstrating that it benefits dense prediction by recovering spatial details and improving boundary fidelity with text guidance. With the refinement module (RM), the model reaches 82.29\% IoU and 84.53\% J@1 (+2.13/+2.26), showing that refinement further enhances grasp reliability and segmentation consistency.
Finally, replacing the static convolution heads with the proposed LDCH results in the best overall performance: 83.14\% IoU and 85.36\% J@1 (+0.85/+0.83), validating that instruction-conditioned kernel mixing provides more robust and task-adaptive predictions.


\subsection{Real-Robot Interactive Grasp Experiments}

\begin{figure*}[t!]         
\centering
  \includegraphics[width=0.95\linewidth]{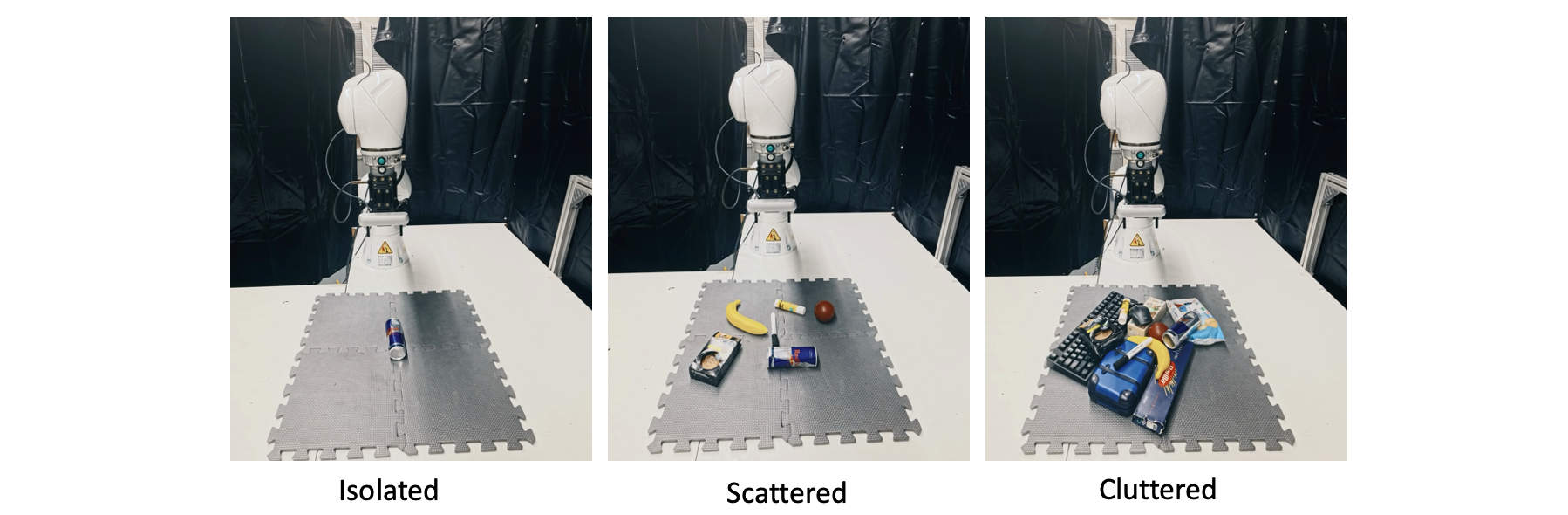}
  \caption{Interactive real robot experiment in 3 Setups: Isolated (left), Scattered (middle), Cluttered (right). }
  \label{fig:real_ex}
\end{figure*}

\begin{figure*}[t!]         
\centering
  \includegraphics[width=0.95\linewidth]{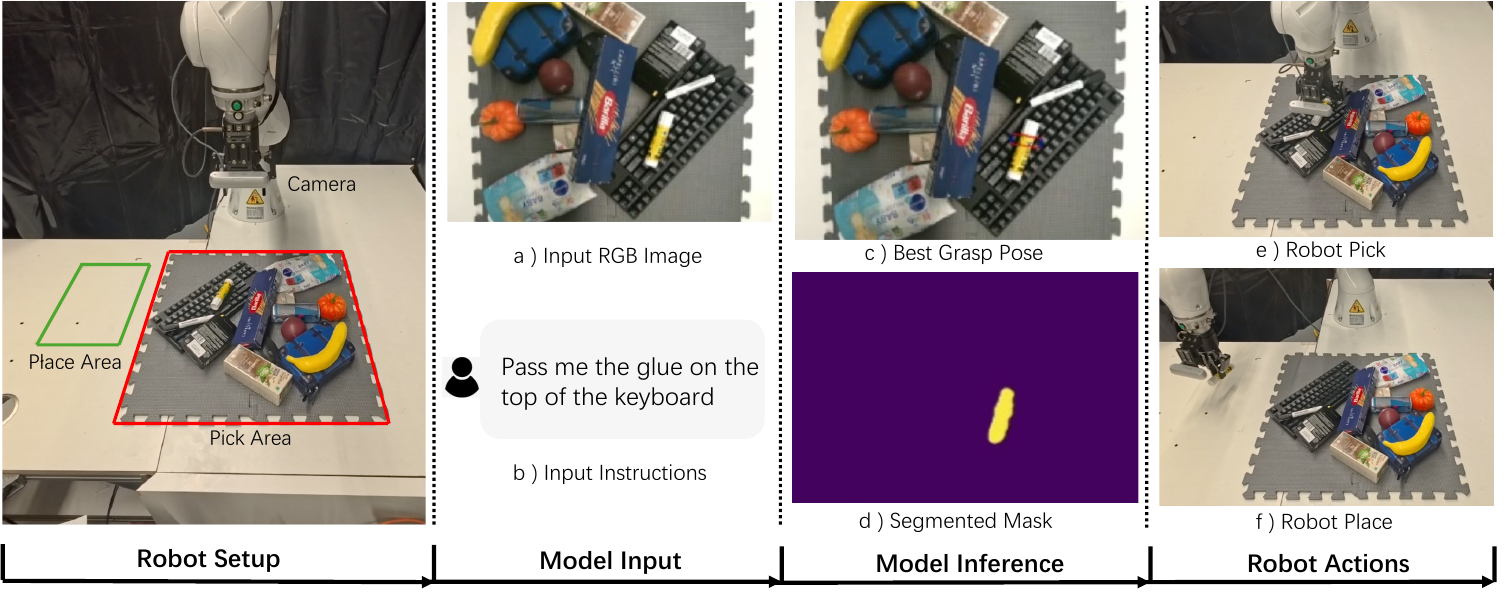}
  \caption{Overview of a real-robot interactive pick-and-place trial. Left: experimental workspace with a wrist-mounted RGB-D camera and predefined pick and place regions. Given an input RGB image (a) and a language instruction (b), the model predicts the best grasp pose (c) and the target segmentation mask (d). The robot then executes the pick (e) and places the object at the predefined destination (f).}
  \label{fig:real_ex_flow}
\end{figure*}

\begin{table}[t!]
\centering
\caption{Results of Real Robot Experiment.}
\label{tab:real_ex}
\begin{tabular}{l|ccccc}
\toprule
\textbf{Setup} & \textbf{Ball} & \textbf{Fruit} & \textbf{Can} & \textbf{Glue} & \textbf{Average}\\
\midrule
Isolated  & 95\%  & 85\%  & 95\%  & 100\%  & 93.75\% \\
Scattered & 100\% & 95\%  & 95\%  & 100\%  & 97.5\% \\
Cluttered & 90\%  & 90\%  & 85\%  & 90\%   & 88.75\% \\
\bottomrule
\end{tabular}
\end{table}

We evaluate the proposed system in an interactive grasping setting, where a user issues an instruction specifying a target object and the robot must localize the correct target and execute a pick-and-place operation. This experiment is designed to assess end-to-end language-conditioned manipulation performance under varying tabletop layouts. We consider four target objects, Ball, fruit, Can, and Glue, as representative items with different shapes. 
Our experiments are conducted on a real robotic platform using a KUKA LBR iiwa 14 R820 manipulator equipped with a Robotiq 2F-85 adaptive gripper. Model inference and the control pipeline are executed on a workstation with an NVIDIA RTX 4090 (24GB) GPU and an AMD Ryzen Threadripper Pro 7955WX CPU.
To assess sensitivity to object arrangement and occlusion, trials are performed under three tabletop configurations:
\begin{enumerate}
    \item \textbf{Isolated}, where the target is presented without nearby distractors;
    \item \textbf{Scattered}, where 4-5 objects are distributed across the workspace with clear separation; 
    \item \textbf{Cluttered}, where objects are densely packed, increasing both occlusion and physical interference.
\end{enumerate}
Representative scenes are shown in Figure~\ref{fig:real_ex}.

Figure~\ref{fig:real_ex_flow} further illustrates the procedure of a representative trial. Specifically, the workspace is partitioned into a predefined pick region and a destination (place) region. For each trial, the system takes an RGB observation together with a user instruction (We use only RGB for inference; depth is used only for execution/calibration.), and outputs an instruction-conditioned target mask and a corresponding grasp pose. These predictions are then used to execute a pick-and-place behavior on the physical robot.
Performance is summarized using the task success rate, computed over repeated executions for each target object and scene configuration. A trial is counted as successful if the robot grasps the instructed target object, achieves a stable pickup, and places it at a predefined destination location. For each target object in each scene configuration, 20 trials are conducted to ensure consistency.
Quantitative results are reported in Table~\ref{tab:real_ex}. In the Isolated setting, success rates are 95\% for Ball, 85\% for Fruit, 95\% for Can, and 100\% for Glue. Under the Scattered configuration, success rates remain high across all targets. In the Cluttered setting, performance decreases to 90\% (Ball), 90\% (fruit), 85\% (Can), and 90\% (Glue).
Across the four targets, the averaged success rate is 93.75\% for Isolated scene, 97.5\% for Scattered scene, and 88.75\% for Cluttered scene. In our real-robot trials, failures are most frequently observed when the arm must execute longer motions across the workspace and when specular highlights caused by lighting introduce reflective artifacts on object surfaces, which can degrade perception and subsequently affect grasp execution. Overall, the system produces stable qualitative behavior on the physical platform, and the observed success rates are consistent with the trends reported earlier in our dataset-based evaluation.



\section{Conclusion}
This work investigates the language-guided robotic grasping in cluttered and ambiguous environments, where robots must accurately localize and grasp objects based on both visual perception and natural language instructions. To this end, we introduced LGGD, an end-to-end grasp detection framework that performs deep cross-modal integration between RGB images and textual queries. Our model distributes linguistic information throughout the entire grasp prediction pipeline, rather than fusing modalities only at an early stage. Through a feature fusion module with dual-path cross-attention, a language-conditioned upsampling mechanism, and a text-guided decoder, our model progressively refines spatial-semantic alignment between language intent and visual features. Furthermore, the residual refinement stage enhances grasp robustness and reduces prediction noise in cluttered scenes. Experimental evaluations on the OCID-VLG and Grasp-Anything++ datasets demonstrate that LGGD significantly outperforms previous language-guided grasping baselines in both segmentation accuracy and grasp success metrics. Real-robot experiments further confirm its practical effectiveness, showing that the model can reliably interpret human instructions and generate feasible grasp poses even in visually complex scenarios. Overall, this research highlights the potential of deeply integrated vision-language reasoning for robotic manipulation. The proposed framework represents a step toward more semantically grounded, instruction-aware robotic systems that can flexibly collaborate with humans in real-world environments.


\bibliographystyle{IEEEtran}
\bibliography{egbib}

\end{document}